\documentclass[runningheads]{llncs}
\usepackage{graphicx}

\usepackage{tikz}
\usepackage{comment}
\usepackage{amsmath,amssymb} 
\usepackage{color}
\usepackage{xspace}
\usepackage{booktabs}
\usepackage{nicefrac}
\usepackage{bbm}
\usepackage{bm}
\usepackage{tabularx,verbatim}
\usepackage{multirow}

\definecolor{remark}{rgb}{1,.5,0} 
\definecolor{citecolor}{rgb}{0,0.443,0.737} 
\definecolor{linkcolor}{rgb}{0.956,0.298,0.235} 
\definecolor{cyan}{rgb}{0.831,0.901,0.945}

\usepackage[accsupp]{axessibility}  

\usepackage[pagebackref,breaklinks,colorlinks,bookmarks=false]{hyperref}
\colorlet{dark-blue}{blue!70!black}
\colorlet{dark-green}{green!80!black}
\colorlet{dark-red}{red!80!black}
\hypersetup{
 colorlinks=true,%
 citecolor=citecolor,%
 filecolor=citecolor,%
 linkcolor=linkcolor,%
 urlcolor=magenta
}
\usepackage{url}

\DeclareMathOperator*{\argmax}{arg\,max}

\begin{document}
\pagestyle{headings}
\mainmatter
\def\ECCVSubNumber{101}  

\makeatletter
\def\blfootnote{\xdef\@thefnmark{}\@footnotetext}
\makeatother

\newcommand{\ours}{FRoST\xspace}
\newcommand{\res}{ResTune\xspace}
\newcommand{\data}{\mathcal{D}}
\newcommand{\lab}{\mathtt{L}}
\newcommand{\unlab}{\mathtt{U}}
\newcommand{\all}{\mathtt{A}}
\newcommand{\old}{\mathtt{old}}
\newcommand{\new}{\mathtt{new}}

\newcommand{\tlab}{\ensuremath{^{[\lab]}}}
\newcommand{\tunlab}{\ensuremath{^{[\unlab]}}}
\newcommand{\tall}{\ensuremath{^{[\all]}}}
\newcommand{\told}{\ensuremath{^{[\old]}}}
\newcommand{\tnew}{\ensuremath{^{[\new]}}}

\newcommand{\task}{\mathcal{T}}
\newcommand{\X}{\mathcal{X}}
\newcommand{\Y}{\mathcal{Y}}
\newcommand{\classes}{\mathcal{C}}
\newcommand{\ie}{\textit{i}.\textit{e}.}
\newcommand{\eg}{\textit{e}.\textit{g}.}
\newcommand{\etal}{\textit{et}.\textit{al}.}
\newcommand{\normal}{\mathcal{N}}
\newcommand{\setting}{class-iNCD\xspace}
\newcommand{\ths}{\textsuperscript{th}\;}
\newcommand{\E}{\mathbb{E}}

\newcommand{\mbf}[1]{\mathbf{#1}}
\newcommand{\va}{\mbf{a}}
\newcommand{\vb}{\mbf{b}}
\newcommand{\vc}{\mbf{c}}
\newcommand{\vd}{\mbf{d}}
\newcommand{\ve}{\mbf{e}}
\newcommand{\vf}{\mbf{f}}
\newcommand{\vg}{\mbf{g}}
\newcommand{\vh}{\mbf{h}}
\newcommand{\vi}{\mbf{i}}
\newcommand{\vj}{\mbf{j}}
\newcommand{\vk}{\mbf{k}}
\newcommand{\vl}{\mbf{l}}
\newcommand{\vm}{\mbf{m}}
\newcommand{\vn}{\mbf{n}}
\newcommand{\vo}{\mbf{o}}
\newcommand{\vp}{\mbf{p}}
\newcommand{\vq}{\mbf{q}}
\newcommand{\vr}{\mbf{r}}
\newcommand{\vu}{\mbf{u}}
\newcommand{\vv}{\mbf{v}}
\newcommand{\vw}{\mbf{w}}
\newcommand{\vx}{\mbf{x}}
\newcommand{\vy}{\mbf{y}}
\newcommand{\vz}{\mbf{z}}
\newcommand{\MA}{\mbf{A}}
\newcommand{\MB}{\mbf{B}}
\newcommand{\MC}{\mbf{C}}
\newcommand{\MD}{\mbf{D}}
\newcommand{\MF}{\mbf{F}}
\newcommand{\MG}{\mbf{G}}
\newcommand{\MH}{\mbf{H}}
\newcommand{\MI}{\mbf{I}}
\newcommand{\MJ}{\mbf{J}}
\newcommand{\MK}{\mbf{K}}
\newcommand{\ML}{\mbf{L}}
\newcommand{\MM}{\mbf{M}}
\newcommand{\MP}{\mbf{P}}
\newcommand{\MQ}{\mbf{Q}}
\newcommand{\MR}{\mbf{R}}
\newcommand{\MS}{\mbf{S}}
\newcommand{\MT}{\mbf{T}}
\newcommand{\MU}{\mbf{U}}
\newcommand{\MV}{\mbf{V}}
\newcommand{\MW}{\mbf{W}}
\newcommand{\MX}{\mbf{X}}
\newcommand{\MY}{\mbf{Y}}
\newcommand{\MZ}{\mbf{Z}}
\newcommand{\ME}{\mbf{E}}

\title{Class-incremental Novel Class Discovery} 

\makeatletter
\newcommand{\printfnsymbol}[1]{%
  \textsuperscript{\@fnsymbol{#1}}%
}
\makeatother

\titlerunning{Class-incremental Novel Class Discovery}
%
\author{Subhankar Roy\inst{1,2} \and
Mingxuan Liu\inst{1} \and
Zhun Zhong\inst{1} \and \\
Nicu Sebe\inst{1} \and
Elisa Ricci\inst{1,2}
}
\authorrunning{S. Roy et al.}
%
\institute{University of Trento, Trento, Italy \and
Fondazione Bruno Kessler, Trento, Italy\\
\email{\{subhankar.roy, mingxuan.liu, zhun.zhong, niculae.sebe, e.ricci\}@unitn.it}}
\maketitle

\begin{abstract}

We study the new task of class-incremental Novel Class Discovery (class-iNCD), which refers to the problem of discovering novel categories in an unlabelled data set by leveraging a pre-trained model that has been trained on a labelled data set containing disjoint yet related categories. Apart from discovering novel classes, we also aim at preserving the ability of the model to recognize previously seen base categories. Inspired by rehearsal-based incremental learning methods, in this paper we propose a novel approach for class-iNCD which prevents forgetting of past information about the base classes by jointly exploiting base class feature prototypes and feature-level knowledge distillation. We also propose a self-training clustering strategy that simultaneously clusters novel categories and trains a joint classifier for both the base and novel classes. This makes our method able to operate in a class-incremental setting. Our experiments, conducted on three common benchmarks, demonstrate that our method significantly outperforms state-of-the-art approaches. Code is available at \url{https://github.com/OatmealLiu/class-iNCD}.

\end{abstract}

\section{Introduction}
\label{sec:intro}

\begin{figure}
    \centering
    \includegraphics[width=\linewidth]{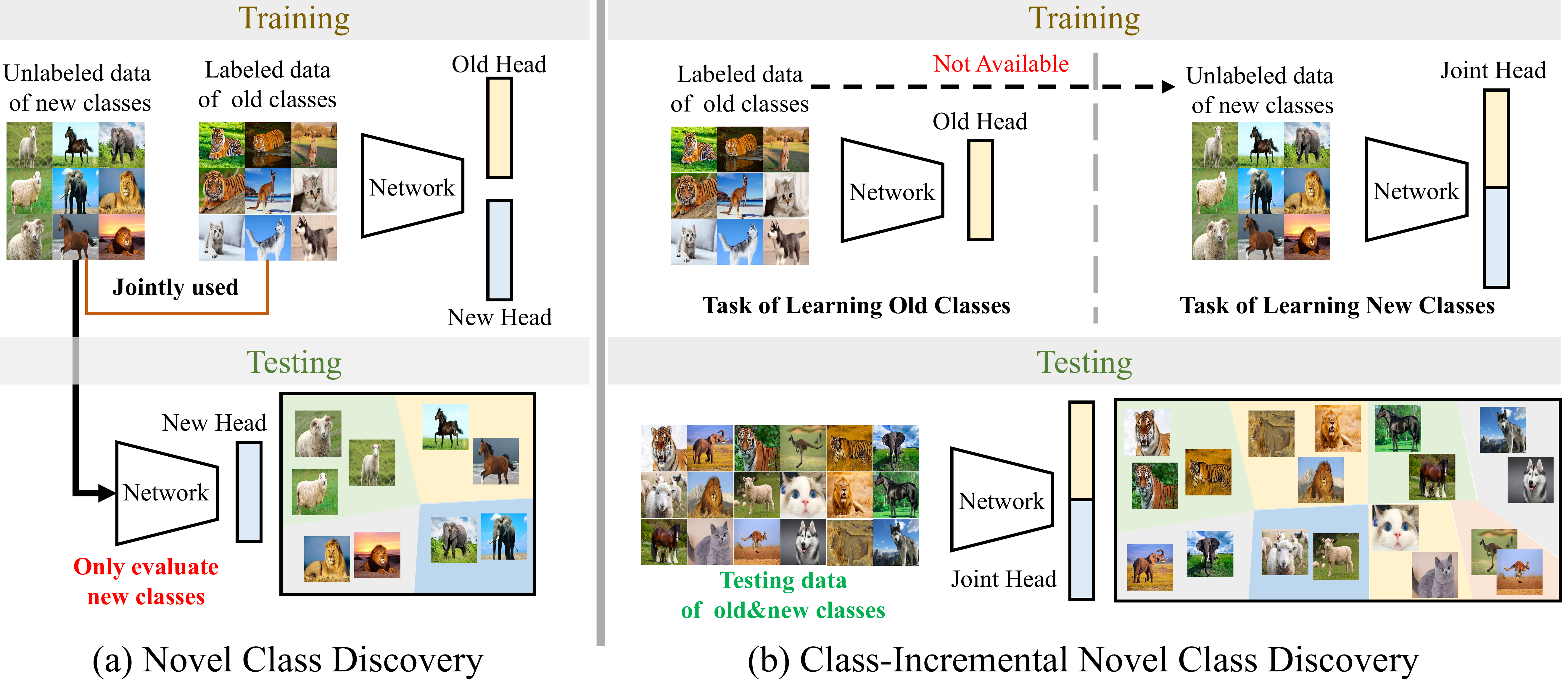}
    \caption{Comparison between the settings (a) Novel Class Discovery (NCD) which solely concerns the performance of novel classes, and (b) the proposed class-incremental NCD (class-iNCD) measures performance of all the classes seen so far with a single classifier.}
    \label{fig:teaser}
\end{figure}

Humans\blfootnote{The first two authors contributed equally.~~Corresponding author: Zhun Zhong} are bestowed with the excellent cognitive skills to learn continually over their lifetime~\cite{french1999catastrophic}, and in most cases without the need of explicit supervision \cite{anderson2015can}. Thus, it has been a long-standing goal of the machine learning research community to build Artificial Intelligence (AI) systems that can mimic this human-level performance. In an attempt to realize this, much effort has been dedicated to learn deep learning models from large reservoirs of both labelled~\cite{krizhevsky2012imagenet,he2016deep,dosovitskiy2020image} and unlabelled data~\cite{caron2020unsupervised,caron2021emerging}. Aside from being effective learners, by imitating human learning mechanisms, neural networks should also be flexible to absorb novel concepts (or \textit{classes}) after having learned some patterns with the past data. The task of automatically discovering novel (or \textit{new}) classes in an unsupervised fashion while leveraging some previously learned knowledge is referred to as \textit{novel class discovery} (NCD)~\cite{han2020automatically,han2019learning,zhong2021neighborhood,zhong2021openmix,fini2021unified} (see Fig.~\ref{fig:teaser}(a)). NCD has gained significant attention in the recent times due to its practicality of efficiently learning novel classes without relying on large quantities of unlabelled data~\cite{han2020automatically}.

Most of the proposed NCD solutions rely on stage-wise~\cite{hsu2017learning,hsu2019multi,han2019learning} or joint \cite{han2020automatically,zhong2021neighborhood,fini2021unified} learning schemes on the labelled and the unlabelled data, with the assumption that structures discovered on the labelled images could be leveraged as a proxy supervision on the unlabelled images. It has been shown that NCD benefits more when the model is trained jointly on the labelled data while using a clustering objective on the unlabelled data~\cite{han2020automatically,zhong2021openmix,zhong2021neighborhood,fini2021unified}. However, access to the labelled data after the pre-training stage can not always be guaranteed in real-world applications due to privacy or storage issues. This calls for a more pragmatic NCD setting where the labelled images would be discarded and only the pre-trained model could be transferred for learning the novel classes. Being meaningful, such \textit{source-free} model adaptation has been explored in the related areas of domain adaptation~\cite{liang2020we,zheng2021rectifying}. Although seems more practical, such a training scheme would gradually cause the network to erase all the previously learned information about the old (or \textit{base}) classes. This drop in the base class performance when the labelled data set becomes unavailable is primarily attributed to the phenomenon of \textit{catastrophic forgetting}~\cite{delange2021continual} in neural networks. In most of the aforementioned NCD methods the performance on the novel classes are only deemed important, without any consideration for preserving the performance on the base classes. We believe that such a setting is of little practical significance in the real world because the adapted model becomes unusable on the base classes and retraining is infeasible.

Given the inherent drawbacks of the existing NCD setting, we argue that an ideal NCD method should aim to learn novel classes without the explicit presence of the labelled data and at the same time preserve the performance on the base classes. This new setting is referred to as \textit{task-incremental} NCD (iNCD), and indeed has been very recently studied in~\cite{liu2022residual}. In details, ResTune~\cite{liu2022residual} uses knowledge distillation~\cite{li2017learning} on the network logits to prevent forgetting on the base classes and a clustering objective \cite{xie2016unsupervised} with task specific network weights for the novel classes. As opposed to the ResTune \cite{liu2022residual}, which facilitates iNCD by solely improving the ability of the network to learn novel classes, we additionally improve the incremental learning aspect in iNCD as well. Specifically, inspired by the rehearsal-based incremental learning methods \cite{buzzega2020dark,chaudhry2019continual,rebuffi2017icarl} which are known to be effective, we propose to store the base class feature \textit{prototypes} from the previous task as exemplars, instead of raw images. Features derived from the stored prototypes are then \textit{replayed} to prevent forgetting old information on the base classes in addition to feature-level knowledge distillation. On the other hand, to facilitate learning of novel classes, we dedicate a task specific classifier that is optimized with robust rank statistics \cite{han2020automatically}. Disadvantageously, the introduction of task specific classifier leads to the dependence on the task-id of an input sample during inference. To overcome reliance on task-id, we propose to maintain a joint classifier for both the base and novel classes, which is trained with the pseudo-labels generated by the task specific one. We call this setting as \textit{class-incremental} NCD (class-iNCD) as it does not allow the task-id information to be used during inference. The high level overview of the new class-iNCD setting is shown in the Fig.~\ref{fig:teaser}(b). As our proposed method amalgamates \textbf{F}eature \textbf{R}eplay and Distillati\textbf{o}n with \textbf{S}elf-\textbf{T}raining we name it \ours. In summary, the contributions of this work are three-fold:
\begin{itemize}
    \item We propose a novel framework, \ours , that can tackle the newly introduced and relevant task of class-incremental novel class discovery (class-iNCD).
    \item Our \ours is equipped with prototypes for feature-replay and employs feature-level knowledge distillation to prevent forgetting. Moreover, 
    it uses pseudo-labels from the task specific head to efficiently learn novel classes without interference, enabling us to achieve a  task-agnostic classifier.
    \item We run extensive experiments on three common benchmarks to prove the effectiveness of our method.  \ours also obtains state-of-the-art performance when compared with the existing baselines. Additionally, we run experiments on a sequence of tasks of unlabelled sets and verify its generality.
\end{itemize}
\section{Related Works}
\label{sec:related}

\noindent \textbf{Novel Class Discovery} (NCD) deals with the task of learning to discover new semantic classes in an unlabelled data set by utilizing the knowledge acquired from another labelled data set~\cite{han2019learning}. It is assumed that the classes in the labelled and unlabelled set are disjoint. So far, several NCD methods have been proposed and they can be broadly classified into two broad sub-categories. The first category of NCD methods use a stage-wise training scheme where the model is first pre-trained on the labelled set, followed by fine-tuning on the unlabelled data using an unsupervised clustering loss~\cite{hsu2017learning,hsu2019multi,han2019learning,liu2022residual}. Barring~\cite{liu2022residual}, none of the above methods consider to tackle the forgetting issue, and as a result the model loses the ability to classify the base classes. The second category comprise of NCD methods that assume both the labelled and unlabelled data are available simultaneously, which are then trained jointly~\cite{han2020automatically,zhong2021neighborhood,zhong2021openmix,jia2021joint,fini2021unified}. As demonstrated in~\cite{liu2022residual}, the NCD methods which rely on joint training always outperform the stage-wise NCD methods. However, the latter family of NCD methods rely on the availability of labelled data, which is often not permitted due to privacy reasons. This makes stage-wise training scheme favourable to tackle class-iNCD, but it lacks the capability to prevent forgetting. Similar to the ResTune~\cite{liu2022residual}, we also build our framework that can be trained in a stage-wise manner and also be able to maintain performance on the base classes. Different from the ResTune we use the predictions of the novel class classifier as pseudo-labels (PL) to train a single joint classifier that can classify both base and novel classes.

\noindent \textbf{Incremental Learning} (IL) is a learning paradigm where a model is trained on a sequence of tasks such that data from only the current task is available for training, while the model is evaluated on all the observed tasks. The IL methods are designed so as to prevent catastrophic forgetting~\cite{goodfellow2013empirical} of the model on the old tasks and at the same time flexible enough to learn on new tasks~\cite{chaudhry2018riemannian}. Most early IL methods addressed the \textit{task-incremental learning} setting (task-IL), where the model has access to a task-id for choosing the task-specific classifier during the testing phase. Given the practical limitations of knowing the task-id during inference, more recent IL methods have started to address the \textit{class-incremental learning} (class-IL) setting, where the task-id is not available during inference. This makes class-IL setting practical and at the same time more challenging than the task-IL setting. Our \ours also operates in the class-IL setting, which we call as class-iNCD. Existing IL methods can be sub-divided into three broad categories: \textit{regularization-based} methods~\cite{kirkpatrick2017overcoming,zenke2017continual,li2017learning,dhar2019learning}, \textit{exemplar-based} methods~\cite{rebuffi2017icarl,castro2018end,buzzega2020dark,chaudhry2019continual} and methods focused on \textit{task-recency bias} problem~\cite{wu2019large}. We refer the readers to the survey in~\cite{masana2020class} for an exhaustive list of class-IL methods. In our \ours we propose to use a combination of knowledge distillation~\cite{li2017learning} at intermediate feature-level and storage of base class feature prototypes as exemplars to prevent forgetting in feature extractor and classifier, respectively. We discuss later in Sec. \ref{sec:method} why this choice is suitable for the class-iNCD setting.

\section{Method}
\label{sec:method}

In this section we describe our \ours for the task of class-iNCD. Before delving into the detail we lay down some preliminaries related to our method.

\noindent \textbf{Problem Definition and Notation}. In the setting of class-incremental novel class discovery (class-iNCD) we are initially given $n\tlab$ instances of a labelled data set $\data\tlab = \{(\vx\tlab_i, \vy\tlab_i)\}^{n\tlab}_{i=1}$ belonging to the supervised task $\task\tlab$, where $\vx\tlab \in \X\tlab$ represents the input images and $\vy\tlab \in \Y\tlab$ as $|\classes\tlab|$-dimensional one-hot labels. Once standard supervised training is finished on the task $\task\tlab$, the data set $\data\tlab$ is discarded and we are presented with $n\tunlab$ instances from a new task $\task\tunlab$. The task $\task\tunlab$ has an unlabelled data set $\data\tunlab = \{\vx\tunlab_j\}^{n\tunlab}_{j=i}$ where $\vx\tunlab \in \X\tunlab$ are the unlabelled images containing $\classes\tunlab$ classes. As in any NCD setting~\cite{han2019learning}, it is assumed that the labels in $\Y\tlab$ and $\Y\tunlab$ are disjoint, \ie, $\Y\tlab \cap \Y\tunlab = \emptyset$. The goal of class-iNCD is to cluster the images in $\data\tunlab$ by just leveraging the learnt information contained in the mapping function $f\tlab \colon \X\tlab \to \Y\tlab$, while still behaving well on the previous task $\task\tlab$. In other words, we are interested in learning a single mapping function $f \colon \X \to \Y\tlab \cup \Y\tunlab$ that can be used to infer the label of any test image $\vx \in \{\X\tlab \cup \X\tunlab\}$. This is in sharp contrast to the existing NCD methods where the performance on $\task\tlab$ is not of interest.

\begin{figure}[!t]
    \centering
    \includegraphics[width=\linewidth]{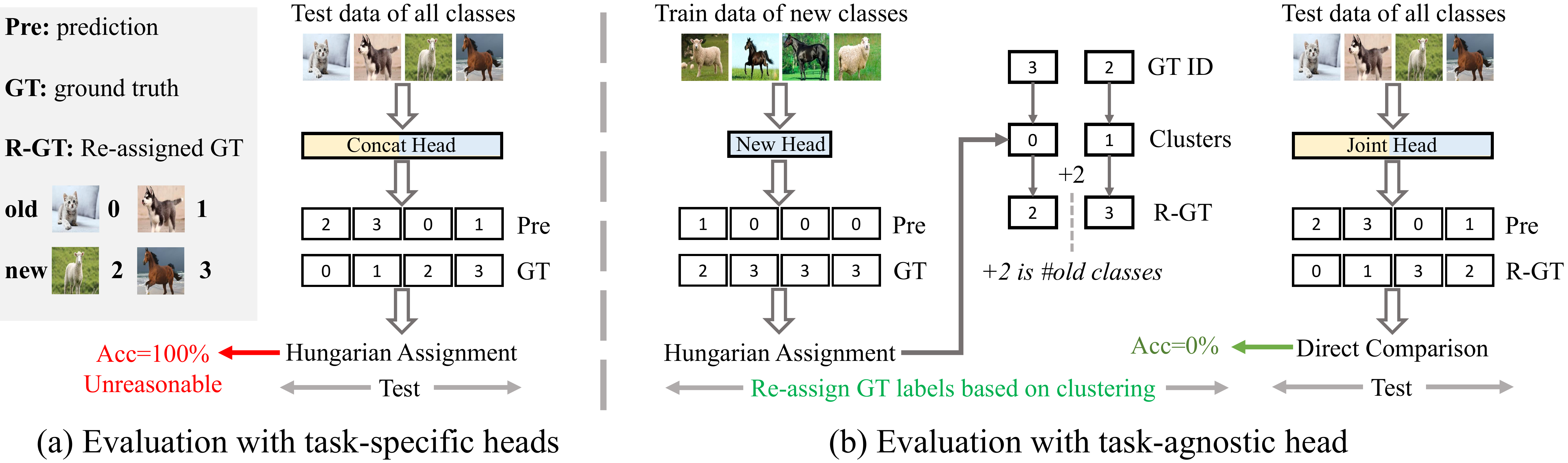}
    \caption{Evaluation protocol comparison (a) evaluation with task-specific heads in iNCD~\cite{liu2022residual} and (b) evaluation with task-agnostic head in our class-iNCD.}
    \label{fig:evaluation}
\end{figure}

\noindent \textbf{Evaluation Protocol}. In the NCD methods~\cite{fini2021unified,zhong2021neighborhood}, task-specific heads are trained for old and new classes\footnote{When referring to classes, we regard old \& base; and, new \& novel interchangeably.}, respectively. This poses a limitation, as they can only operate in task-specific NCD setting. To address this problem, ResTune~\cite{liu2022residual} uses the concatenation of old and new heads during inference. The class-incremental performance is estimated with the Hungarian Assignment (HA)~\cite{kuhn1955hungarian} by regarding this problem as a clustering task. However, this evaluation protocol is indeed improper in class-iNCD, since it does not explicitly distinguish the old and new classes. As shown in the Fig.~\ref{fig:evaluation}(a), the classifier recognizes the samples of old classes as novel classes (and vice versa), and yet the accuracy obtained by HA is still 100\%, making the evaluation in~\cite{liu2022residual} unfair.

In this work, we learn a task-agnostic head (or \textit{joint} head) and propose a new evaluation protocol for class-iNCD (see Fig.~\ref{fig:evaluation}(b)). In details, we first use the \textit{new head} to estimate the predictions of unlabeled data from the new classes. We utilize the HA~\cite{kuhn1955hungarian} to re-assign ground-truth IDs based on the predictions and ground-truth labels for the new classes only. The joint (task-agnostic) classifier is used to evaluate the new classes test samples by directly comparing the predictions with these re-assigned ground-truth labels. Whereas for the old classes test data, we evaluate using the old classes ground truth. As shown in Fig.~\ref{fig:evaluation}(b), our evaluation protocol explicitly distinguishes the old and new classes. As evident, our evaluation is more reasonable than~\cite{liu2022residual} and penalizes the metric when the new classes are classified as one of the old classes, which is an ideal behaviour.

\begin{figure}[!t]
    \centering
    \includegraphics[width=\linewidth]{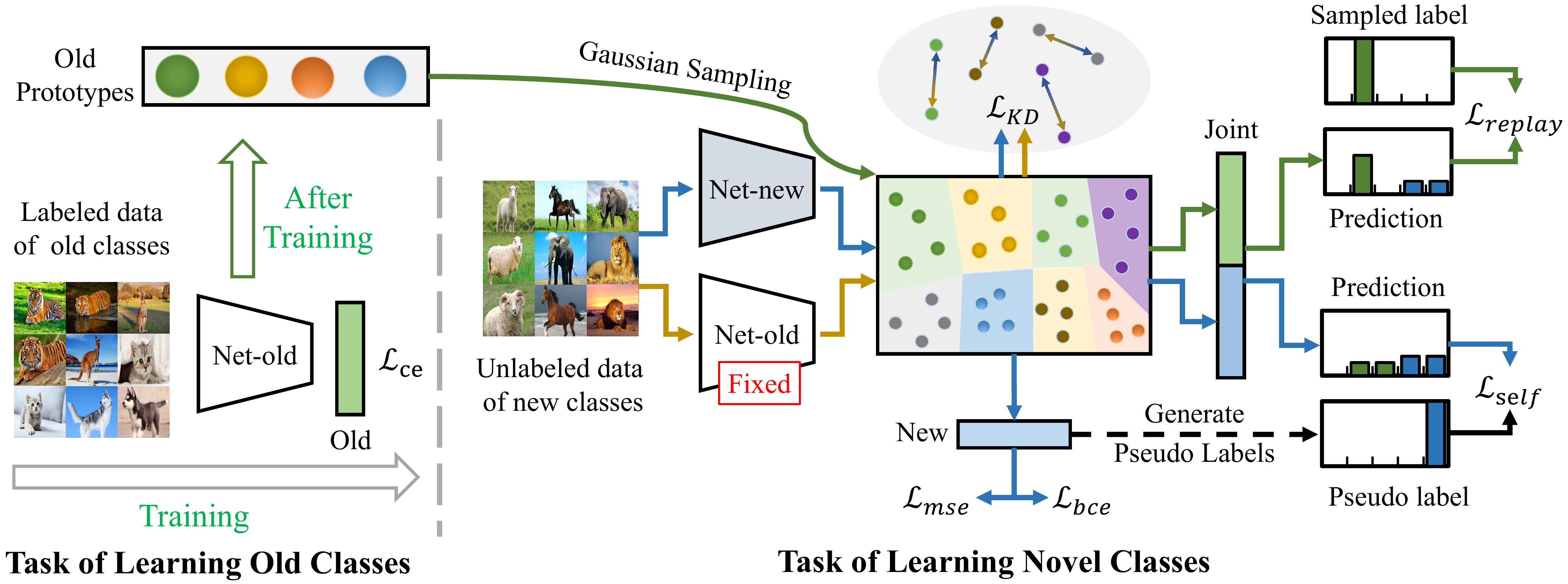}
    \caption{An overview of the proposed FroST. \textbf{Left}: a base model is learned supervisedly ($\mathcal{L}_\text{ce}$) on the old classes. Old class-prototypes and variances are stored. \textbf{Right}: the new classes are learned with a clustering objective ($\mathcal{L}_\text{bce}$). Forgetting on old classes is prevented by using feature-distillation ($\mathcal{L}_\text{KD}$) and feature-replay ($\mathcal{L}_\text{replay}$) with the class-prototypes. A joint classifier is learned by self-training ($\mathcal{L}_\text{self}$) with pseudo-labels.}
    \label{fig:framework}
\end{figure}

\noindent \textbf{Overall Framework}. Being in the incremental learning setting, our proposed \ours (see Fig.~\ref{fig:framework}) operates in two stages. In the first stage we learn the mapping function $f\tlab \colon \X\tlab \to \Y\tlab$ in a supervised manner on the labelled data set $\data\tlab$ that can recognize samples belonging to the first $\classes\tlab$ categories. We model the function $f\tlab$ with a neural network that is further composed of two sub-networks: feature extractor $g(\cdot)$ and a linear classifier $h\tlab(\cdot)$ that outputs $\classes\tlab$ logits, such that $f\tlab = h\tlab \circ g$. The feature extractor $g$ and classifier $h\tlab$ are parameterized by $\theta_g$ and $\theta_{h\tlab}$, respectively. Before we move to the second stage, we compute per-class intermediate feature prototypes $\bm{\mu}_c$  from the intermediate features $\vz\tlab =g(\vx\tlab)$, belonging to each class $c$. Additionally, we also compute and store the variance of the features of class $c$ as ${\bm{v}_c}^2$.

In the second stage, the $\data\tlab$ is discarded and the novel classes are learned on $\data\tunlab$ by reusing the transferred network weights $f\tlab$. Since our goal is to learn an unique classifier that can accommodate $\classes\tall = \classes\tlab + \classes\tunlab$ classes, we extend the classifier $h\tlab$ to $h\tall$ in order to incorporate the $\classes\tunlab$ novel classes. Besides $h\tall$, we instantiate a new task-specific classifier $h\tunlab$ for $\task\tunlab$ that is trained on $\data\tunlab$ to exclusively classify the novel classes. The classifiers $h\tall$ and $h\tunlab$ are parameterized by $\theta_{h\tunlab}$ and $\theta_{h\tall}$, respectively. In details, the network $f\tunlab = h\tunlab \circ g$ is trained using the clustering objective in~\cite{han2020automatically} that leverages previously learned information to provide supervision using the robust rank statistics. With the goal of learning a joint classifier $h\tall$, we obtain pseudo-label for $\vx\tunlab$ from $h\tunlab$ and distill it to the newly extended part of $h\tall$, which handles the novel classes. On the other hand, to mitigate forgetting on the base classes of $\task\tlab$ we employ two strategies: \textit{feature-level} knowledge distillation~\cite{hinton2015distilling,li2017learning} on $g$ that ensures the feature encoding for the old task $\task\tlab$ does not drift too far while learning on $\task\tunlab$; and \textit{generative feature-replay} drawn from a Gaussian distribution $\normal(\bm{\mu}_c, {\bm{v}_c}^2)$ is used to preserve performance of the top part of the $h\tall$, which is responsible for classifying the base classes. During inference the classifier $h\tall$ is used.

\subsection{Preliminaries}
\label{sec:prelim}

\textbf{Supervised Training}. In the first stage of the \setting task we have at disposal the labelled images from $\data\tlab$. This stage consists in learning a supervised model $f\tlab$ that can classify the base classes drawn from the task $\task\tlab$. We aim to learn the parameters (\{$\theta_g, \theta_{h\tlab}$\}) of the model $f\tlab = h\tlab \circ g$ by using a supervised \textit{cross-entropy} loss:

\begin{equation}
\label{eqn:ce_loss}
    \mathcal{L}_\mathrm{ce} = - \E_{p(\vx\tlab, \vy\tlab)} \frac{1}{C\tlab}\sum^{C\tlab}_{k=1} y\tlab_{k} \log \sigma_k (h\tlab (g(\vx\tlab))),
\end{equation}
where $\sigma_k(\vl) = \nicefrac{\exp(l_k)}{\sum_j \exp(l_j)}$ represents the likelihood corresponding to the $k$\ths output from the model and $C\tlab$ is the number of classes in the task $\task\tlab$.

\noindent\textbf{Knowledge Distillation to Prevent Forgetting}. Having learned an optimal model on a given task, the main challenge in IL is to learn new tasks without forgetting the past information. A very popular regularization-based approach to overcome forgetting on previously learned tasks is by using \textit{knowledge distillation} (KD)~\cite{hinton2015distilling}. Concretely, based on KD, \textit{Learning without Forgetting} (LwF)~\cite{li2017learning} is an effective method commonly used in IL. It consists in penalizing the network if the representation of data from previous tasks drifts too far while learning on a new task. Assuming a simplified task-IL learning scenario containing just two tasks: $\task\told$ and $\task\tnew$, where a model $f\told = h\told \circ g$ has already been trained using the objective in Eq.~(\ref{eqn:ce_loss}) and a new task $\task\tnew$ has been presented to the learning algorithm. The goal of LwF is to prevent forgetting on $\task\told$ while learning on $\task\tnew$. LwF  keeps a copy of the old model $f\told = h\told \circ g\told$ and simultaneously creates a new instance $f\tnew = h\tnew \circ g\tnew$ (with $g\tnew$ = g\told) for learning on $\task\tnew$. The $f\tnew$ differs from $f\told$ in the final classification head where the task-specific classifier $h\tnew$ exclusively handles the class assignment for the new classes in $\task\tnew$. Given a sample $\vx\tnew$ from the new task $\task\tnew$, LwF aims to match the pre-recorded logits $\va\told = h\told(g\told(\vx\tnew))$ from the frozen $f\told$ with the old task logits $\hat{\va}\told = h\told(g\tnew(\vx\tnew))$. Essentially, this prevents $g\tnew$ to produce feature encoding that is too different from that of $g\told$, since the success for the old task heavily depends on it. Formally, the LwF loss at logits-level is given as:
\begin{equation}
\label{eqn:lwf}
    \mathcal{L}^{\mathrm{logits}}_{\mathrm{KD}} = - \E_{p(\vx\tnew)} \frac{1}{K\told}\sum^{K\told}_{k=1} \pi_{k} (h\told(g\told(\vx\tnew))) \log \pi_{k} (h\told(g\tnew(\vx\tnew))),
\end{equation}
where $\pi_k(\va) = \nicefrac{\exp(a_k / \tau)}{\sum_j \exp(a_j / \tau)}$ is the temperature controlled likelihood of the model with $\tau$ being the temperature. The parameters $(\{\theta_{g\tnew}, \theta_{h\told}\})$ corresponding to $g\tnew$ and $h\told$ are updated with Eq.~(\ref{eqn:lwf}). However, the need of having separate task-specific classifiers in the LwF approach limits the applicability of such models to the task-IL setting, as in ResTune \cite{liu2022residual}. While LwF can ideally be extended to the class-IL consisting of a single classifier, it would require pre-allocation of all the logits during the first task. The assumption of knowing apriori the cardinality of the tasks and their contituent classes is impractical. Thus, we build on top LwF and adapt it to the \setting. 

\subsection{Class-incremental Novel Class Discovery}
\label{sec:proincd}

We are interested in learning a model that can incrementally cluster unlabelled images into a set of novel classes, after the model has been trained on a labelled set of images. Besides good performance on the novel classes we also desire to preserve the performance on the previously seen classes, without having access to or storing images from the previous tasks. Most importantly, at any point of time during the training sessions, we maintain a single classification head for all the classes seen so far. To address the challenging task of \setting we propose to tackle it from two different axes. The first axis is concerned with learning discriminative features on the unlabelled data set $\data\tunlab$ by using a clustering objective. Although the model gets better at classifying the novel classes, it's performance gradually deteriorates on the base classes due to forgetting~\cite{delange2021continual}. To overcome this issue, the second axis deals with preventing forgetting on all base classes by using the images only from the new task, combined with a feature-replay strategy. We elaborate them below.

\noindent \textbf{Self-training for Novel Class Discovery}. When presented with an unlabelled data set $\data\tunlab$, the discovery step in class-iNCD involves learning the weights of the network $f\tunlab = h\tunlab \circ g$. While the newly initialized classifier $h\tunlab$ yet lacks the capability to classify images into novel categories, the feature extractor $g$ on the other hand has already been trained on a related labelled data set $\data\tlab$ and has a notion of what constitutes a semantic concept in an image. Adopting this ideology from the NCD method AutoNovel~\cite{han2020automatically}, the pairwise similarity between a pair of unlabelled images ($\vx\tunlab_i$, $\vx\tunlab_j$) is inferred and provided as a weak form of supervision in the discovery step. The feature descriptors $\vz\tunlab_i = g(\vx\tunlab_i)$ and $\vz\tunlab_j = g(\vx\tunlab_j)$ corresponding to the pair ($\vx\tunlab_i$, $\vx\tunlab_j$) are then compared using the robust rank statistics. If the top-$k$ ranked dimensions of the feature descriptor pair ($\vz\tunlab_i, \vz\tunlab_j$) are found the same then ($\vx\tunlab_i$, $\vx\tunlab_j$) can be considered to belong to the same class. The pairwise pseudo-label is formulated as:

\begin{equation}
\label{eqn:ranking}
    \tilde{y}\tunlab_{ij} = \mathbbm{1}\{\mathrm{top}_k(\vz\tunlab_i) = \mathrm{top}_k(\vz\tunlab_j)\},
\end{equation}
where $\mathrm{top}_k \colon \vz\tunlab \to \mathcal{S}\{(1, \dots, k)\} \subset \mathcal{P}\{(1, \dots, |\vz\tunlab|)\}$ denotes the subset of top-$k$ most activated feature indices in $\vz\tunlab$. This pairwise pseudo-label is then used to train the classifier $h\tunlab$ for the novel classes. In detail, the dot-product of the classifier's predictions $p_{ij} = \sigma(\langle h\tunlab(g(\vx\tunlab_i)), h\tunlab(g(\vx\tunlab_j))\rangle)$ can be interpreted as a similarity between $\vx\tunlab_i$ and $\vx\tunlab_j$, where $\sigma(\cdot)$ is a logistic function. Thus, the pairwise pseudo-label $\tilde{y}\tunlab_{ij}$ computed in Eq.~(\ref{eqn:ranking}) is used to enforce this association between $\vx\tunlab_i$ and $\vx\tunlab_j$. The parameters (\{$\theta_g, \theta_{h\tunlab}$\}) are trained with a \textit{binary cross-entropy} loss as:
\begin{equation}
\label{eqn:bce}
    \mathcal{L}_\mathrm{bce} = - \E_{p(\vz\tunlab)} \tilde{y}\tunlab_{ij} \log (p_{ij}) + (1 - \tilde{y}\tunlab_{ij}) \log (1 - p_{ij}).
\end{equation}

While the objective in Eq.~(\ref{eqn:bce}) learns a classifier for the novel classes, such training scheme makes the inference step dependent on task-id like ResTune. In order to make our model suitable for \setting we resort to self-training with the help of pseudo-labels that are computed from $f\tunlab$ to train the joint classifier $h\tall$. In details, given the goal of learning the model $f\tall = h\tall \circ g$, we use $h\tunlab$ to compute the pseudo-label $\hat{y}\tunlab$ for an unlabelled image $\vx\tunlab$.  The $\hat{y}\tunlab$ is then used to supervise the training of $h\tall$. The self-training loss is described as:
\begin{equation}
\label{eqn:self-train}
    \mathcal{L}_\mathrm{self} = - \E_{(\vx\tunlab, \hat{\vy}\tunlab)} \frac{1}{|C\tall|} \sum^{|C\tall|}_{k=1} \hat{y}\tunlab_k \log \sigma_k (h\tall(g(\vx\tunlab))),
\end{equation}
where
\begin{equation}
    \hat{y}\tunlab = C\tlab + \argmax_{k \in C\tunlab} h\tunlab(g(\vx\tunlab)).
\end{equation}

Since the pairwise pseudo-labels in Eq.~(\ref{eqn:ranking}) can be noisy, it can lead to a poorly trained $h\tunlab$. As a consequence, the noisy pseudo-labels $\hat{y}\tunlab$ from $h\tunlab$ can have an adverse impact on the training of the joint classifier $h\tall$. To minimize the cascading error propagation we also enforce consistency between two correlated views for an unlabelled image $\vx\tunlab$. Specifically, using stochastic data augmentation on $\vx\tunlab$ we generate a correlated view $\bar{\vx}\tunlab$ and optimize (\{$\theta_g, \theta_{h\tunlab}$\}) with a \textit{mean-squared error} loss as:
\begin{equation}
\setlength{\abovedisplayskip}{1.5pt}
\setlength{\belowdisplayskip}{1.5pt}
\label{eqn:mse}
    \mathcal{L}_\mathrm{mse} = \E_{p(\vx\tunlab, \bar{\vx}\tunlab)} \frac{1}{|C\tunlab|} \sum^{|C\tunlab|}_{k=1} \Big( \sigma_k \big(h\tunlab(g(\vx\tunlab))\big) - \sigma_k \big(h\tunlab(g(\bar{\vx}\tunlab))\big)\Big)^2.
\end{equation}

Finally, the overall loss for discovering novel classes and having a single classifier for all the classes seen so far can be written as:
\begin{equation}
\label{eqn:ncd}
    \mathcal{L}_\mathrm{novel} = \mathcal{L}_\mathrm{bce} +  \omega_\mathrm{self}(t) \mathcal{L}_\mathrm{self} +  \omega_\mathrm{mse}(t) \mathcal{L}_\mathrm{mse},
\end{equation}
where $\omega_\mathrm{self}(t)$ and $\omega_\mathrm{mse}(t)$ are ramp-up functions to ensure stability in learning.

\noindent \textbf{Feature Replay and Distillation for Class-incremental Learning}. While the proposed self-training assists the model $f\tall$ in discovering the novel classes, it simultaneously loses the ability to predict the old classes in $\task\tunlab$. To mitigate the forgetting we propose feature replay and feature distillation. To recall, at the end supervised training on $\task\tlab$ and before discarding $\data\tlab$ we compute the class prototype $\bm{\mu}\tlab_c$ and variance ${\bm{v}\tlab_c}^2$ for each base class as:
\begin{equation}
\label{eqn:prototypes}
    \bm{\mu}\tlab_c = \frac{1}{n\tlab_c} \sum^{n\tlab_c}_{i=1}  g(\vx\tlab_i),  \quad \quad {\bm{v}\tlab_c}^2 = \frac{1}{n\tlab_c} \sum^{n\tlab_c}_{i=1} (g(\vx\tlab_i) - \bm{\mu}\tlab_c)^2,
\end{equation}
where $n\tlab_c$ denotes the number of samples belonging to class $c$ in $\data\tlab$. While learning on the new task $\task\tunlab$; the weights of the joint classifier $h\tall$, corresponding to the base classes $C\tlab$, are trained by replaying features from the class-specific Gaussian distribution $\mathcal{N}(\bm{\mu}\tlab_c, {\bm{v}\tlab_c}^2)$ of $\task\tlab$. The feature-replay loss is given as:
\begin{equation}
\label{eqn:replay}
    \mathcal{L}_\mathrm{replay} = - \E_{c \sim C\tlab} \E_{(\vz\tlab, \vy\tlab_c) \sim \mathcal{N}(\bm{\mu}_c, {\bm{v}_c}^2)} \sum^{|C\tall|}_{k=1} y\tlab_{kc} \log \sigma_k(h\tall(\vz\tlab)).
\end{equation}

As the feature extractor $g$ also gets updated during the optimization of Eq.~(\ref{eqn:ncd}), this will make the prototypes outdated. To keep the feature replay useful we add an extra regularization on $g$ with the help of feature distillation, which is given as:
\begin{equation}
\label{eqn:feat-KD}
    \mathcal{L}^{\mathrm{feat}}_{\mathrm{KD}} = - \E_{p(\vx\tunlab)} \Big|\Big|g\tlab(\vx\tunlab) - g(\vx\tunlab)\Big|\Big|_2,
\end{equation}
where $g\tlab$ is the feature extractor from the previous task and is kept frozen. 

Conventionally, in supervised class-IL or task-IL, regularization with the LwF loss in the logits space while learning supervisedly on a new task is usually effective in preventing forgetting. Contrarily in \setting, as the novel classes need to be learned without explicit supervision, it makes the optimization of NCD part interfere with that of forgetting. This motivates us to decouple the objective for \textit{not-forgetting} into $\mathcal{L}_\mathrm{replay}$ and $\mathcal{L}^{\mathrm{feat}}_{\mathrm{KD}}$. We show later in Sec. \ref{sec:exp} with adequate experiments the disadvantages of using LwF on the logits of the network. The overall objective for not-forgetting past information is given as:
\begin{equation}
\label{eqn:not-forget}
    \mathcal{L}_\mathrm{past} = \mathcal{L}_\mathrm{replay} + \lambda \mathcal{L}^{\mathrm{feat}}_{\mathrm{KD}},
\end{equation}
where $\lambda$ is used to weight the feature distillation loss. 

\noindent \textbf{Overall Training}. Finally, our  \ours is optimized with the following objective:
\begin{equation}
\label{eqn:frost}
    \mathcal{L}_\mathrm{\ours} = \mathcal{L}_\mathrm{novel} + \mathcal{L}_{\mathrm{past}}.
\end{equation}

\section{Experiments}
\label{sec:exp}

\subsection{Experimental Setup}
\label{sec:exp-setup}

\noindent \textbf{Datasets}. We have used three data sets to conduct experiments for class-iNCD: CIFAR-10~\cite{krizhevsky2009learning}, CIFAR-100~\cite{krizhevsky2009learning} and Tiny-ImageNet~\cite{le2015tiny}. We split the data sets into the old and new classes following the existing NCD and iNCD works~\cite{han2020automatically,zhong2021neighborhood,liu2022residual}. The splits are reported in the supplementary material.

\noindent\textbf{Evaluation metrics}. We used our new evaluation protocol (Sec.~\ref{sec:method}) to evaluate the performance on the test data for all the classes. We report three classification accuracies, denoted as \textbf{Old}, \textbf{New} and \textbf{All}. They represent the accuracy obtained from the joint classifier head on the samples of the old, new and old+new classes, respectively. Refer to the supplementary material for details.

\noindent \textbf{Implementation details} We used ResNet-18~\cite{he2016deep} as the backbone in all the experiments. We have adopted most of the hyperparameters from AutoNovel~\cite{han2020automatically}. We introduce only one additional hyperparemeter $\lambda$, which is set to 10. Rest of the implementation details can be found in the supplementary material.

\subsection{Ablation Studies}
\label{sec:ablations}

\begin{figure}[!t]
    \centering
    \includegraphics[width=0.8\linewidth]{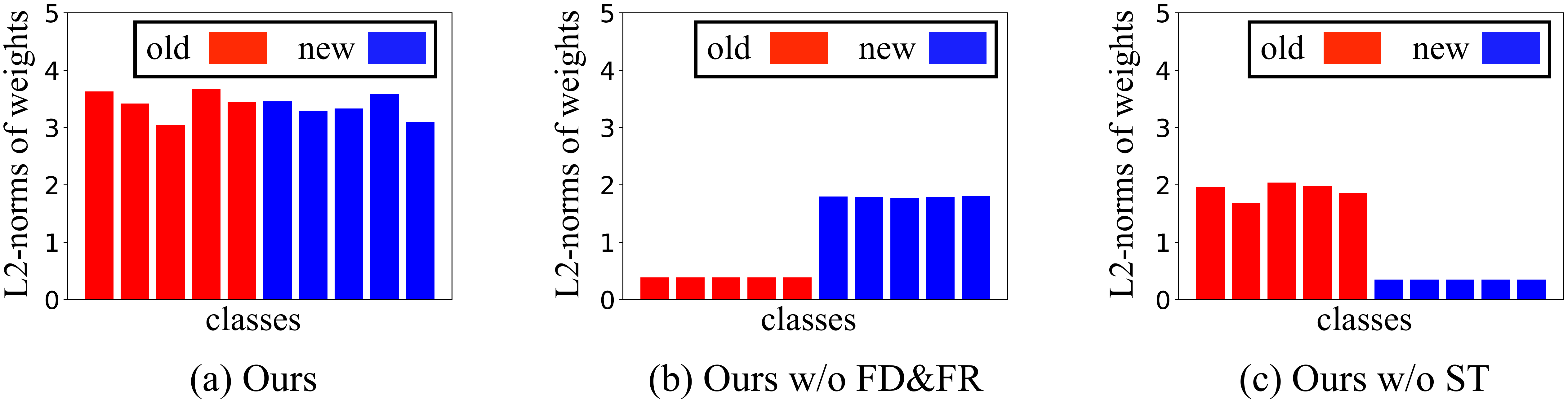}
    \caption{Comparison of L2 norms of the classifier weights. Our full method has balanced L2-norms leading to a better balance in classification for old and new classes.}
    \label{fig:weight-norm}
\end{figure}

\noindent\textbf{Effectiveness of Feature Replay and Distillation}. 
In Tab. \ref{tab:ablation_loss} we measure the impact of the components proposed for not forgetting: feature distillation (FD), and feature replay (FR). The \ours without FD and FR results in complete forgetting of the old classes. This is not surprising because without FD the feature extractor has moved far away from the original configuration. Moreover, as the joint classifier weights corresponding to the new classes are only optimized during the NCD stage (due to the disabled loss Eq.~(\ref{eqn:replay})), it leads to what is called the task-recency bias, resulting in higher norms for the new classes weights (see Fig.~\ref{fig:weight-norm}). In other words, for any old test sample, the classifier is highly biased to predict the new classes, leading to complete misclassification of old classes. Similar effect is observed when FD is enabled but FR is disabled.

\begin{table}[!t]
\scriptsize
\centering
    \caption{Ablation study of the proposed feature distillation (FD), feature replay (FR) and self-training (ST) that form our \ours for the \setting.}

    \begin{tabular}{l|ccc|ccc|ccc|ccc}
        \toprule
         \multirow{2}{*}{Methods} & \multicolumn{3}{c|}{CIFAR-10} & \multicolumn{3}{c|}{CIFAR-100} & \multicolumn{3}{c|}{Tiny-ImageNet} & \multicolumn{3}{c}{Average}\\ 
          &  Old & New &  All & Old & New &  All & Old & New &  All & Old & New & All \\ 
         
         \hline
         \ours (Ours) & 77.4 & 49.5 & \textbf{63.5} & 62.5 & 45.8 & \textbf{59.2} & 54.4 & 33.9 & \textbf{52.4} & 64.8 & 43.1 & \textbf{58.3} \\
         
         w/o FD \& FR & 0.0 & 36.4 & 18.2 & 0.0 & 33.1 & 6.6 & 0.0 & 37.2 & 3.7 & 0.0 & 35.6 & 9.5\\
         
         w/o FD & 0.0 & 39.4 & 19.7 & 0.0 & 33.1 & 6.6 & 0.0 & 34.3 & 3.4 & 0.0 & 35.6 & 9.9\\
         
         w/o FR & 0.0 &\textbf{ 73.3} & 36.6 & 0.0 & \textbf{57.8} & 11.6 & 0.0 & \textbf{40.9} & 4.1 & 0.0 & \textbf{57.3} & 17.4 \\
         \hline
         w/o ST & \textbf{91.7} & 0.0 & 45.8 & \textbf{69.2} & 0.0 & 55.4 & \textbf{57.5} & 0.0 & 51.7 & \textbf{72.8} & 0.0 & 51.0\\
         w/o FD \& FR \& ST & 16.6 & 0.0 & 8.3 & 2.7 & 0.0 & 2.1 & 2.0 & 0.0 & 1.8 & 7.1 & 0.0 & 4.1\\
         \bottomrule
    \end{tabular}
    \label{tab:ablation_loss}
\end{table}

\begin{figure}[!t]
    \centering
    \includegraphics[width=0.95\linewidth]{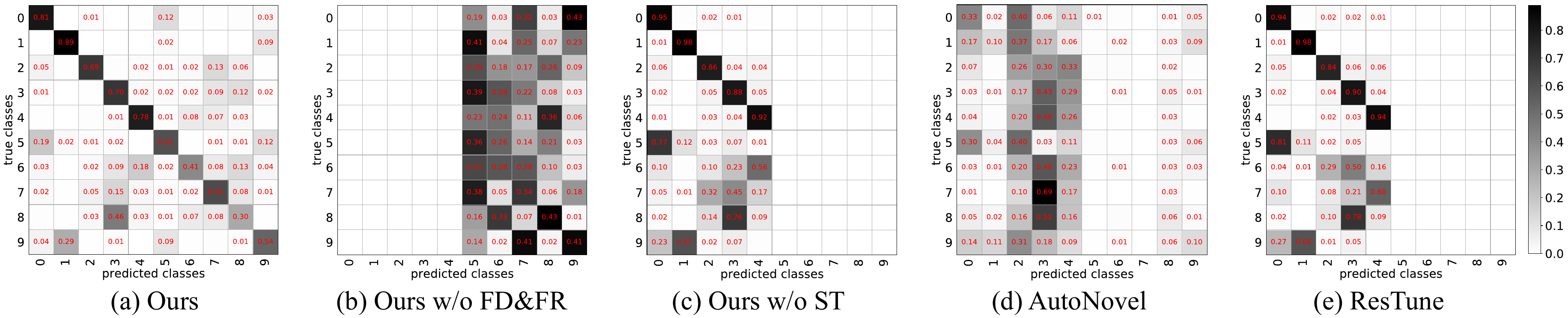}
    \caption{Comparisons of confusion matrix of different methods. Note that, the label IDs of novel classes are re-assigned by our evaluation protocol.}
    \label{fig:confusion}
\end{figure}

\noindent\textbf{Effectiveness of Self-Training}. In the bottom half of Tab. \ref{tab:ablation_loss} we show the impact of the absence of self-training (ST) on the performance. The \ours w/o ST causes no interference in the optimization from the new classes and the joint classifier is able to preserve the performance on old classes. This highlights the truly complex nature of balancing the performance of both old and new classes in the \setting setting. This phenomenon is visualized in Fig.~\ref{fig:weight-norm} through the norms of the weights of the joint classifier where exists large discrepancies between the old versus new classes. Similar conclusions can also be drawn by observing the confusion matrix reported in Fig. \ref{fig:confusion}. Furthermore, when we drop FD and FR along with ST, we notice a further degradation of the performance of old classes, demonstrating the positive impact of FR and FD in not forgetting. 

\noindent\textbf{Comparison of our Feature Replay and Distillation with LwF}. Here we empirically demonstrate the need of decoupling the LwF objective into FR and FD when it comes to learning a joint classifier for the \setting. As a control experiment, we use the original formulation of LwF (as in Eq.~(\ref{eqn:lwf})) as a drop-in replacement for our FR and FD. We optimize with the original LwF loss, applied both at the softmax and pre-softmax outputs, and found that compared to \ours the performance on the new classes improves but at the cost of a large drop in the old classes performance. We conjecture that since the weights for the new classes are randomly initialized in $h\tall$ at the start of the NCD stage, the joint classifier as a whole learns at a different rate than the feature extractor $g$, which is already pre-trained on the old classes. As the LwF loss optimizes both the $g$ and $h\tall$, it causes slow-fast learning interference from the new classes. This is evident from better new classes performance \textit{w.r.t} \ours. As shown in Tab. \ref{tab:ablation_LwF}, indeed adding FR to LwF improves the performance on the old classes (\textit{e.g.}, 6.8\% vs 49.9\% for Tiny-ImageNet). This again proves the effectiveness of FR and the need of decoupled \textit{not-forgetting} objective in the \setting.

\begin{table}[!ht]
\centering
\scriptsize
    \caption{Ablation study comparing \ours with LwF (logits-KD).}

{
    \begin{tabular}{c|ccc|ccc|ccc|ccc}
        \toprule
         \multirow{2}{*}{IL Methods} & \multicolumn{3}{c|}{CIFAR-10} & \multicolumn{3}{c|}{CIFAR-100} & \multicolumn{3}{c|}{Tiny-ImageNet} & \multicolumn{3}{c}{Average}\\ 
          & Old & New &  All & Old & New &  All & Old & New &  All & Old & New & All \\ 
         \hline
         \ours (Ours) & \textbf{77.4} & 49.5 & \textbf{63.5} & \textbf{62.5} & 45.8 & \textbf{59.2} & \textbf{54.4} & 33.9 & \textbf{52.4} & \textbf{64.8} & 43.1 & \textbf{58.3} \\
         
         \hline
         LwF (softmax) & 13.6 & 63.2 & 38.4 & 7.4 & \textbf{63.5} & 18.6 & 2.1 & \textbf{42.8} & 6.2 & 7.7 & 56.5 & 21.1 \\
         LwF (softmax) + FR & 21.4 & 61.1 & 41.3 & 33.3 & 61.2 & 38.9 & 35.3 & 33.1 & 35.0 & 30.0 & 51.8 & 38.4 \\
         
         LwF (pre-softmax) & 19.4 & 76.3 & 47.9 & 13.6 & 61.4 & 23.2 & 6.8 & 38.7 & 10.0 & 13.3 & \textbf{58.8} & 27.0 \\
         
         LwF (pre-softmax) + FR & 24.8 & \textbf{77.5} & 51.1 & 49.3 & 58.3 & 51.1 & 49.9 & 26.8 & 47.6 & 41.3 & 54.2 & 49.9 \\
         \bottomrule
    \end{tabular}
}
\label{tab:ablation_LwF}
\end{table}

\noindent\textbf{Effect of using Joint and Novel Classifiers}. Here we elaborate on the choice of using the joint $h\tall$ and novel $h\tunlab$ classification heads in \ours to address \setting. We report the $\texttt{Joint}$ baseline in Tab. \ref{tab:ablation_head}, where we discard the novel classifier head and solely use the joint classifier for learning the new classes and the old classes. We find that this approach causes hindrance in learning the new classes for CIFAR10 and Tiny-ImagenNet data sets because two parts of the same classifier are subject to gradients of different magnitudes, highlighting the need to decouple the learning of two tasks. In the next ablation, we also disable the ST with pseudo-labels that are generated by joint itself and we find that it destabilizes the performance on the new classes. Finally, we construct an ablation where we do not extend $h\tlab$ to $h\tall$, but instead use $h\tlab$ in conjunction with $h\tunlab$ and is denoted with $\texttt{Novel w/o ST}$. We observe that this behaves similarly with the previous ablation analysis of $\texttt{\ours w/o ST}$ in Tab.~\ref{tab:ablation_loss}. Thus, we conclude that having joint and novel heads trained with ST is crucial in \setting.

\begin{table}[!t]
\centering
\scriptsize
\caption{Ablation study on having a single and separated heads for old and new classes. Joint: class-agnostic head; Novel: new classes classifier head.}
{
    \begin{tabular}{c|ccc|ccc|ccc|ccc}
        \toprule
         \multirow{2}{*}{Classifier Head} & \multicolumn{3}{c|}{CIFAR-10} & \multicolumn{3}{c|}{CIFAR-100} & \multicolumn{3}{c|}{Tiny-ImageNet} & \multicolumn{3}{c}{Average} \\ 
         
          & Old & New &  All & Old & New &  All & Old & New &  All & Old & New & All \\ 
         
         \hline
          
         Joint + Novel (Ours) & 77.4 & \textbf{49.5} & \textbf{63.5} & 62.5 & 45.8 & 59.2 & 54.4 & \textbf{33.9} & \textbf{52.4} & 64.8 & \textbf{43.1} & \textbf{58.3} \\
         
         \hline
         
         Joint  & 81.3 & 41.5 & 61.4 & 64.5 & \textbf{46.3} & \textbf{60.9} & 56.8 & 8.4 & 52.0 & 67.5 & 32.1 & 58.1 \\
         
         Joint w/o ST & 91.7 & 0.0 & 45.8 & 68.6 & 29.4 & 60.7 & 57.5 & 0.1 & 51.7 & \textbf{72.6} & 9.9 & 52.8 \\
         
         Novel w/o ST & \textbf{92.0} & 0.0 & 46.0 & \textbf{67.9} & 32.1 & 60.7 & \textbf{57.9} & 0.0 & 52.1 & \textbf{72.6} & 10.7 & 52.9 \\
         \bottomrule
    \end{tabular}
}
    \label{tab:ablation_head}
\end{table}

\begin{table}[!t]
\scriptsize	
\centering
\caption{Comparison with state-of-the-art methods in class-iNCD.}
    \begin{tabular}{c|ccc|ccc|ccc|ccc}
        \toprule
         \multirow{2}{*}{Methods} & \multicolumn{3}{c|}{CIFAR-10} & \multicolumn{3}{c|}{CIFAR-100} & \multicolumn{3}{c|}{Tiny-ImageNet} & \multicolumn{3}{c}{Average} \\
         
         & Old & New & All & Old & New & All &Old & New & All &Old & New & All  \\ 
         
         \hline
         AutoNovel\cite{han2020automatically} & 27.5 & 3.5 & 15.5 & 2.6 & 15.2 & 5.1 & 2.0 & 26.4 & 4.5 & 10.7 & 15.0 & 8.4 \\
         
         ResTune\cite{liu2022residual} & 91.7 & 0.0 & 45.9 & \textbf{73.8} & 0.0 & 59.0 & 44.3 & 0.0 & 39.9 & \textbf{69.9} & 0.0 & 48.3 \\
         
         NCL\cite{zhong2021neighborhood} & \textbf{92.0} & 1.1 & 46.5 & 73.6 & 10.1 & \textbf{60.9} & 0.8 & 6.5 & 1.4 & 55.5 & 5.9 & 36.3 \\
         
         DTC\cite{Han2019LearningTD} & 64.0 & 0.0 & 32.0 & 55.9 & 0.0 & 44.7 & 35.5 & 0.0 & 32.0 & 51.8 & 0.0 & 36.2 \\
         \hline
         \textbf{FRoST} & 77.5 & \textbf{49.5} & \textbf{63.4} & 64.6 & \textbf{45.8} & 59.2 & \textbf{54.5} & \textbf{33.7} & \textbf{52.3} & 65.5 & \textbf{39.8} & \textbf{54.9} \\
         \bottomrule
    \end{tabular}
\label{tab:benchmark_acc}
\end{table}

\subsection{Comparison with State-of-the-art Methods}
\label{sec:sota}

We compare our \ours with the state-of-the-art NCD methods under the newly proposed class-iNCD setting. We also compare with ResTune~\cite{liu2022residual} which is a recently proposed method for iNCD. As none of these existing methods have been evaluated in the \setting setting, we re-run the baselines and simply modify the evaluation protocol which is described in Sec. \ref{sec:method}. We report the results of the NCD~\cite{han2020automatically,zhong2021neighborhood,han2019learning}, iNCD~\cite{liu2022residual} baselines and \ours in the Tab. \ref{tab:benchmark_acc}. As can be observed, under the \setting all the NCD \cite{han2020automatically,zhong2021neighborhood,han2019learning} fail to obtain a good balance on the old and new classes. Interestingly, while none of these NCD methods use any explicit objectives to prevent forgetting, they tend to predict well the old classes (see column \textbf{Old} in Tab. \ref{tab:benchmark_acc}) and poor performance on new classes (see column \textbf{New} in Tab. \ref{tab:benchmark_acc}). When visualizing the confusion matrix in Fig.~\ref{fig:confusion} we found that most of the test samples get classified as old classes due to the old classes classifier having higher norms. As a consequence, this gives the impression that the baselines methods are able to retain performance on old classes. Second, for the above methods, although the new classes performance obtained with the joint head appears to be low, the actual performance of their novel head in the task-aware evaluation is indeed high. We report the breakdown of the novel classes performance in Tab. \ref{tab:multi-incd} where, for instance, the column \textbf{New-1-N} denotes the task-aware clustering performance of the novel head on the new classes. As can be observed, the new classes classifier of the NCD baselines can indeed learn on the new classes (\textit{e.g.}, 34.2\% in NCL vs 32.4\% in \ours). 

ResTune, although designed specifically for the iNCD setting, exhibits similar counter-intuitive behaviour with the performance on the old classes dominating the new classes. To investigate this pathology, we inspect into the confusion matrix in Fig. \ref{fig:confusion} (e) and find that all the samples get predicted to the first five old classes for CIFAR10. In other words, the overall performance reported in ResTune \cite{liu2022residual} is actually dominated by the old classes performance. We report confusion matrices on bigger data sets in the supplementary material. This shows that the existing evaluation method for iNCD is flawed and our proposed \setting is indeed more meaningful that properly evaluates the effectiveness of a learning algorithm. Contrarily our proposed \ours consistently achieves a good balance in performance in all the tested data sets. This also demonstrates the validity of the components in our proposed \ours. We present a detailed comparison analysis between ResTune and \ours in the supplementary material.

\begin{table}[!t]
\scriptsize	
\caption{Comparison with the state-of-the-art methods in the two-step class-iNCD setting where new classes arrive in two episodes, instead of one. New-1-J: new classes performance from joint head at first step, New-1-N: new classes performance from novel head at first step, etc.}
\begin{center}
    \begin{tabular}{c|cccc|cccccc}
        \toprule
         \multirow{3}{*}{Methods}& \multicolumn{10}{c}{Tiny-ImageNet} \\
         \cline{2-11}
         & \multicolumn{4}{c|}{First Step (180-10)} & \multicolumn{6}{c}{Second Step (180-10-10)} \\
         & Old & New-1-J & New-1-N & All & Old & New-1-J & New-2-J & New-1-N & New-2-N & All \\ 
         \hline
         ResTune\cite{liu2022residual} & 39.7 & 0.0 & 38.0 & 37.6 & 34.9 & 0.0 & 0.0 & 25.4 & 42.8 & 31.4 \\
         
         DTC\cite{Han2019LearningTD} & 38.9 & 0.0 & \textbf{43.8} & 36.9 & 33.4 & 0.0 & 0.0 & 28.0 & \textbf{59.4} & 30.1 \\

         NCL\cite{zhong2021neighborhood} & 5.6 & 0.0 & 34.2 & 5.3 & 1.4 & 0.0 & 2.6 & 21.6 & 41.6 & 1.4 \\
         
         \hline
         \textbf{FRoST} & \textbf{55.2} & \textbf{27.6} & 32.0 & \textbf{53.8} & \textbf{42.5} & \textbf{34.8} & \textbf{31.2} & \textbf{31.2} & 46.8 & \textbf{41.6} \\
         \bottomrule
    \end{tabular}
\end{center}
    \label{tab:multi-incd}
\end{table}

\noindent\textbf{Two-Step Class-iNCD}. As done in the class-IL literature~\cite{masana2020class}, we also run experiments on a sequence of novel tasks, which we call as two-step \setting, where 20 novel classes in Tiny-ImageNet are added in two steps, each step dealing with 10 novel classes. We compare our \ours with the baseline methods in Tab. \ref{tab:multi-incd} where we show not only the joint classifier head performance (\eg, \textbf{New-1-J}), but also from the novel classifier head (\eg, \textbf{New-1-N} and \textbf{New-2-N}) at each step. As can be seen, for the baseline methods the novel classifier heads can satisfactorily discover the new classes at each step, but when evaluated with the joint head biases the predictions to the old classes. Unlike the baselines, \ours does not suffer from this issue and leads to more balanced predictions.

\section{Conclusion}

In this work we addressed the novel problem of class-incremental NCD. This task differs from the traditional NCD since we are not only interested in discovering novel classes but also aim to prevent forgetting on the old classes. To prevent this forgetting phenomenon we proposed feature replay and feature-level distillation that is well suited for the \setting. Moreover, to make inference task-agnostic, we propose to maintain a joint classifier that can classify any of the previously seen classes. We train this joint classifier using the pseudo-labels generated by the novel classifier head that is trained with a clustering loss. We compared our method to many relevant works and obtained superior performance on various benchmarks. Given the practical nature of \setting and encouraging results with our \ours, we believe this work will stimulate further research.

\noindent\textbf{Acknowledgements.} We thank the funding agencies: EU H2020 projects SPRING (No. 871245) and AI4Media (No. 951911); and the EUREGIO project OLIVER.

%
%
\bibliographystyle{splncs04}
\bibliography{egbib}

\clearpage
\appendix
{\centering\Large\bf
 Supplementary Material: Class-incremental Novel Class Discovery}

\hfill
\setcounter{table}{0}
\renewcommand{\thetable}{A\arabic{table}}%
\setcounter{figure}{0}
\renewcommand{\thefigure}{A\arabic{figure}}%
\setcounter{equation}{0}
\renewcommand{\theequation}{A\arabic{equation}}%

\newcommand{\bluebox}{\protect\tikz[baseline=-.7ex]{\protect\node[rectangle,draw=blue,thick] (r) at (0, 0){};}}

This supplementary material is organised as follows: In Sec.~\ref{sec:app-exp-setup} we report additional details about the experimental set-up. In Sec. \ref{sec:app-comparison} we dissect the evaluation protocols for existing incremental NCD method, and compare the performance of ResTune \cite{liu2022residual} with our proposed \ours. Finally, in Sec. \ref{sec:app-two-step} we report additional experiments on CIFAR100 for the two-step \setting setting.

\section{Experimental Set-up}
\label{sec:app-exp-setup}

\noindent\textbf{Datasets}. In the Tab.~\ref{tabel:data-details} we report the standard splits between the old and the new classes for the three benchmarks. Note that for the CIFAR100 and Tiny-ImageNet there is an imbalance between the old and the new classes.

\begin{table}[!h]
  \label{tabel:data-details}
    \caption{Dataset statistics for class-incremental novel class discovery.}
  \scriptsize
  \centering
  \begin{tabularx}{0.88\linewidth}{l|cc|cc|cc}
    \toprule
    \multirow{2}{*}{Dataset}   &  \multicolumn{2}{c|}{Labeled Set (Old classes)}   & \multicolumn{2}{c|}{Unlabeled Set (New classes)} & \multicolumn{2}{c}{Test Set} \\
               &  \#image & \#class   & \#image & \#class  & \#image & \#class  \\
    \hline
    CIFAR-10 & 25K & 5 & 25K & 5 & 5K & 10\\
    CIFAR-100 & 40K & 80 & 10K & 20 & 5K & 100\\
    Tiny-ImageNet & 90K & 180 & 10K & 20 & 10K & 200\\
    \bottomrule
  \end{tabularx}
\end{table}

\noindent\textbf{Implementation Details.} We have trained our model with the SGD optimizer and the initial learning rate set to 0.1, which is then decayed by a factor of 10 after 170 epochs. The total training epoch is 200 and the batch size is set to 128. For the mean-squared error loss, following \cite{han2020automatically} we adopt the ramp-up function with weight $\gamma=\{5, 50, 50\}$ and ramp-up length $T=\{50, 150, 150\}$ for CIFAR-10, CIFAR-100 and Tiny-ImageNet, respectively. For the self-training loss, we use the same ramp-up length of corresponding data set, but use the weight $\gamma=0.05$ for all data sets.

\section{Comparison with ResTune}
\label{sec:app-comparison}

\subsection{ResTune in the Original Setting versus \setting}
\label{sec:app-restune}

In this work we argue that an incremental NCD algorithm should be evaluated in a task-agnostic fashion with a joint classifier (see Sec. 3 and Fig. 2 (b) in the main paper), such that, at any stage in the lifetime of the model, the predicted classes should fall in the corresponding bucket of class indices seen in a given training session. In other words, if the model sees samples from task $\task\told$ containing classes of indices 0-4 in a given stage of training, then at any future inference stage, say after having trained on $\task\tnew$, the model must assign test samples from $\task\told$ to the first five logits. This particular evaluation protocol has been introduced as the \setting setting in our work. While ResTune (RT)~\cite{liu2022residual} has been proposed for the task of incremental NCD, its evaluation differs from our \setting. Concretely, \res reports three evaluation accuracy in their work: {\em (i)} the task-aware accuracy on the old classes (abbreviated as $\text{\textbf{Old}}_\text{RT}$) that uses the ``old'' classifier head; {\em (ii)} the task-aware accuracy on the new classes (abbreviated as $\text{\textbf{New}}_\text{RT}$) that uses the ``new'' classifier head; and {\em (iii)} the task-agnostic clustering accuracy on all classes (abbreviated as $\text{\textbf{All}}_\text{RT}$) that employs the concatenated ``old'' and ``new'' classifier heads (or concat-head). As discussed in Sec. 3 of the main paper, the evaluation using the Hungarian Assignment on all the test data set to obtain $\text{\bf{All}}_\text{RT}$ is unreasonable because some of the new classes may get assigned to old classes, making the inference inconsistent between tasks. We denote the evaluation of \res as $Original_\text{RT}$ protocol. 

\begin{table}[!t]
    \centering
    \scriptsize
    \setlength{\tabcolsep}{8pt}
    \caption{Comparison of the evaluation protocols alongside the classifier heads used in $Original_\text{RT}$ and our proposed \setting. The metrics and the corresponding classifier heads used to obtain them are color coded}
    \begin{tabular}{c|c|c}
        \toprule
         \multirow{2}{*}{Metric} &  \multicolumn{2}{c}{Classifier head} \\
         \cline{2-3}
         & $Original_\text{RT}$ Protocol & \setting Protocol \\
         \midrule
         \textcolor{red}{$\text{\textbf{Old}}_\text{RT}$}/\textcolor{blue}{\textbf{Old}} & \textcolor{red}{old-head} & \textcolor{blue}{joint-head}  \\
         
         \textcolor{red}{$\text{\textbf{New}}_\text{RT}$}/\textcolor{blue}{\textbf{New}} & \textcolor{red}{new-head} & \textcolor{blue}{joint-head}  \\
         
         \textcolor{red}{$\text{\textbf{All}}_\text{RT}$}/\textcolor{blue}{\textbf{All}} & \textcolor{red}{concat-head} & \textcolor{blue}{joint-head}  \\
         \bottomrule
    \end{tabular}
    \label{tab:eval-protocol}
\end{table}

On the other hand, for the \setting setting, we use a single classification head to report three accuracies: \textbf{Old}, \textbf{New} and \textbf{All} corresponding to the samples from the old, new and all the classes, as described in the Sec. 4.1 of the main paper. One crucial difference between the $Original_\text{RT}$ and \setting is that in \setting evaluation protocol we always use a joint classifier head (or the concat-head for \res) to evaluate all the metrics. Moreover, our \setting also uses Hungarian Assignment for the \textbf{All} metric, but only for obtaining the re-assigned ground truth for the ``new'' classes (being unsupervised) and \textit{not} on all the data set (see Fig. 2). This ensures that the samples from the old classes and the new classes are evaluated correctly and rightfully results in a drop in the performance if cross-task class assignment occurs (see Fig. 2 (b)), which should be the desired behaviour for any incremental predictor. We visually demonstrate the difference between the $Original_\text{RT}$ and \setting protocols in Tab. \ref{tab:eval-protocol}. Note than when \res is evaluated in the \setting setting, the concat-head is used because \res by construction has two separate heads.

\begin{table}[!ht]
\centering
\scriptsize
    \caption{The comparison of the \res with \ours using the $Original_\text{RT}$ and the \setting evaluation protocols on the CIFAR-10 data set (5 old classes and 5 new classes). While the \res fairs well in the $Original_\text{RT}$ setting, the new classes performance is dramatically low when tested under the \setting. Contrarily, \ours maintains a balanced performance over all the classes by consistently outperforming \res}
{
    \begin{tabular}{c|ccc|ccc}
        \toprule
        Dataset & \multicolumn{6}{c}{CIFAR-10 (\#Old:5; \#New:5)} \\ 
        \hline
         Protocol & \multicolumn{3}{c|}{$Original_\text{RT}$} &
         \multicolumn{3}{c}{\setting} \\
         \hline
         Method & $\text{\textbf{Old}}_\text{RT}$ & $\text{\textbf{New}}_\text{RT}$ & $\text{\textbf{All}}_\text{RT}$ & \textbf{Old} & \textbf{New} & \textbf{All}\\ 
         \hline
         ResTune\cite{liu2022residual} (in paper) & 85.5 & 89.0 & 52.1 & - & - & - \\
         ResTune\cite{liu2022residual} (reproduce) & 91.7 & 76.7 & 46.9 & \textbf{91.7} & 0.0 & 45.9\\
         \hline
         FRoST (Ours) & - & - & - & 77.5 & \textbf{49.5} & \textbf{63.4}\\
         \bottomrule
    \end{tabular}
}
\label{tab:restune_comparison_cifar10}
\end{table}

\begin{table}[!ht]
\centering
\scriptsize
    \caption{The comparison of the \res with \ours using the $Original_\text{RT}$ and the \setting evaluation protocols on the CIFAR-100 data set (80 old classes and 20 new classes). While the \res fairs well in the $Original_\text{RT}$ setting, the new classes performance is dramatically low when tested under the \setting. Contrarily, \ours maintains a balanced performance over all the classes by consistently outperforming \res}
{
    \begin{tabular}{c|ccc|ccc}
        \toprule
        Dataset & \multicolumn{6}{c}{CIFAR-100 (\#Old:80; \#New:20)} \\ 
        \hline
         Protocol & \multicolumn{3}{c|}{$Original_\text{RT}$} &
         \multicolumn{3}{c}{\setting} \\
         \hline
         Method & $\text{\textbf{Old}}_\text{RT}$ & $\text{\textbf{New}}_\text{RT}$ & $\text{\textbf{All}}_\text{RT}$ & \textbf{Old} & \textbf{New} & \textbf{All}\\ 
         \hline
         ResTune\cite{liu2022residual} (in paper) & 73.8 & 63.7 & 59.1 & - & - & - \\
         ResTune\cite{liu2022residual} (reproduce) & 73.8 & 56.0 & 59.0 & \textbf{73.8} & 0.0 & 59.0 \\
         \hline
         FRoST (Ours) & - & - & - & 64.6 & \textbf{45.8} & \textbf{59.2} \\
         \bottomrule
    \end{tabular}
}
\label{tab:restune_comparison_cifar100}
\end{table}

\begin{table}[!ht]
\centering
\scriptsize
    \caption{The comparison of the \res with \ours using the $Original_\text{RT}$ and the \setting evaluation protocols on the Tiny-ImageNet data set (180 old classes and 20 new classes). While the \res fairs well in the $Original_\text{RT}$ setting, the new classes performance is dramatically low when tested under the \setting. Contrarily, \ours maintains a balanced performance over all the classes by consistently outperforming \res}
{
    \begin{tabular}{c|ccc|ccc}
        \toprule
        Dataset & \multicolumn{6}{c}{TinyImageNet (\#Old:180; \#New:20)} \\ 
        \hline
         Protocol & \multicolumn{3}{c|}{$Original_\text{RT}$} &
         \multicolumn{3}{c}{\setting} \\
         \hline
         Method & $\text{\textbf{Old}}_\text{RT}$ & $\text{\textbf{New}}_\text{RT}$ & $\text{\textbf{All}}_\text{RT}$ & \textbf{Old} & \textbf{New} & \textbf{All}\\ 
         \hline
         ResTune\cite{liu2022residual} (in paper) & 58.0 & 46.3 & 41.2 & - & - & -\\
         ResTune\cite{liu2022residual} (reproduce) & 44.3 & 27.3 & 40.4 & 44.3 & 0.0 & 39.9 \\
         \hline
         FRoST (Ours) & - & - & - & \textbf{54.5} & \textbf{33.7} & \textbf{52.3}\\
         \bottomrule
    \end{tabular}
}
\label{tab:restune_comparison_tiny}
\end{table}

\begin{figure}[!h]
    \centering
    \includegraphics[width=0.9\linewidth]{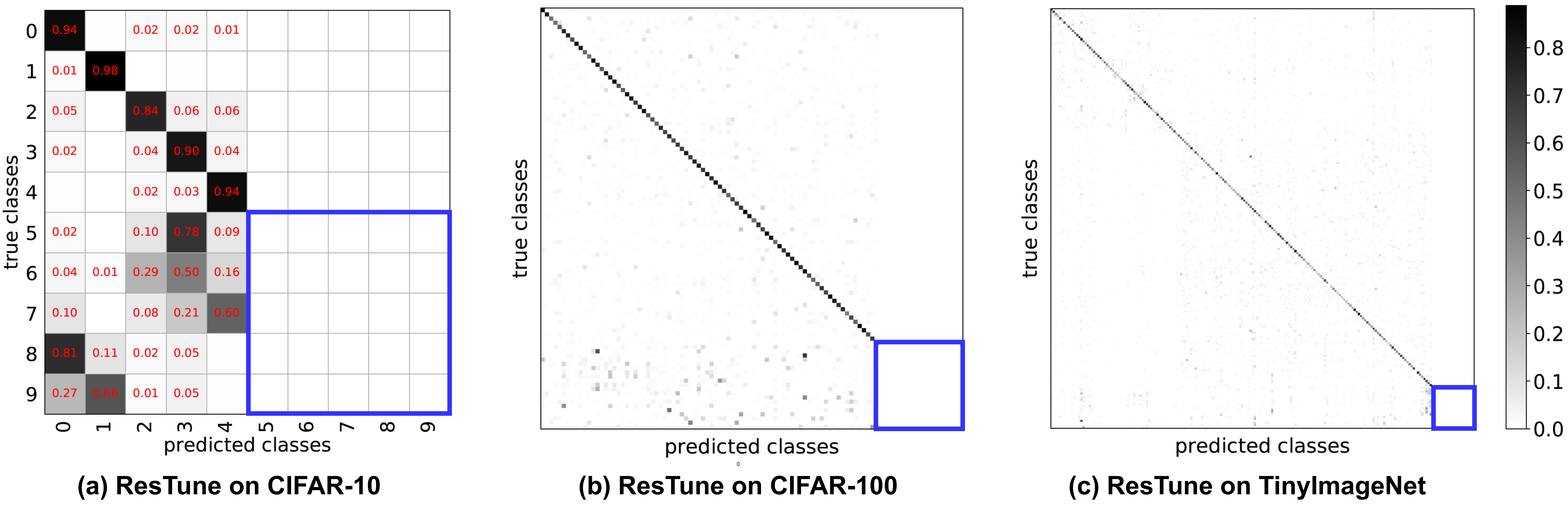}
    \caption{The confusion matrices for \res when evaluated in the task-agnostic case (or $\text{\textbf{All}}_\text{RT}$) of the $Original_\text{RT}$ setting. The \bluebox\; denotes the part of the confusion matrix which corresponds to the new classes. The \res always predicts test samples from the new classes as belonging to the old classes. Digital zoom is recommended}
    \label{fig:cm-restune-settings}
\end{figure}

Next we show that the $Original_\text{RT}$ protocol introduced in \cite{li2017learning} is flawed and gives a false sense of improvement in performance over the erstwhile baselines, given the split chosen between labelled and unlabelled classes. To this end, we re-run \res, using the official code published by the authors\footnote{\url{https://github.com/liuyudut/ResTune}} of \cite{liu2022residual}, to obtain the performance in the $Original_\text{RT}$ setting. We report the numbers of \res in the left halves of the Tab. \ref{tab:restune_comparison_cifar10}, \ref{tab:restune_comparison_cifar100} and \ref{tab:restune_comparison_tiny} under the $Original_\text{RT}$ setting for CIFAR-10, CIFAR-100 and Tiny-ImageNet, respectively. Except for the CIFAR-10, the \res numbers are mostly reproducible for the metric $\text{\textbf{All}}_\text{RT}$, which is of interest to us since it is obtained with the concat-head without the need of task-id. To better understand the distribution of predictions for \res in the task-agnostic classification case we plot the confusion matrix (CM) for all the data sets in Fig. \ref{fig:cm-restune-settings}. We can immediately notice from the CMs that all the samples in the data set have been predicted as one of the old classes. This signifies that the \res simply can not predict any new classes correctly when evaluated in the task-agnostic setting (\ie, $\text{\textbf{All}}_\text{RT}$), indicated by the \textit{empty} blue box in Fig. \ref{fig:cm-restune-settings} (a)-(c). The illusion of performance for the $\text{\textbf{All}}_\text{RT}$ comes from the old classes because cardinality of old classes dominate the new classes (\eg, 80 old classes vs 20 new classes for CIFAR-100, etc). While the $\text{\textbf{New}}_\text{RT}$ is reasonably good, it is task-id dependant and thus, meaningless in practical applications. Therefore, the $Original_\text{RT}$ protocol introduced in \cite{liu2022residual} does not truly reflect the classification capability of a incremental NCD algorithm.

Given, the inherent flaws of the evaluation protocol of $Original_\text{RT}$, in our work we propose the \setting and then evaluate \res using this protocol. In the \setting, the concat-head is always used for evaluating the \res, and the performance is reported in the right halves of the Tab. \ref{tab:restune_comparison_cifar10}, \ref{tab:restune_comparison_cifar100}, \ref{tab:restune_comparison_tiny}. We notice that the \textbf{All} metric for \res do not vary much from the $\text{\textbf{All}}_\text{RT}$. However, there is an acute drop in performance for the \textbf{New} metric in comparison to the $\text{\textbf{New}}_\text{RT}$, dropping all the way to 0\% for all the data sets. As shown previously in Fig. \ref{fig:cm-restune-settings}, this is a consequence of the \res always getting activated in the old logits given any test sample from the new classes. This is expected because in \setting, just like the \textbf{All} metric, the \textbf{New} metric is also task-agnostic. Thus, it can be concluded that the protocol of our \setting can accurately evaluate if an incremental learner is well-behaved for both the new and old classes simultaneously.

\subsection{Comparison of \ours with ResTune}
\label{sec:app-ours}
\begin{figure}[!t]
    \centering
    \includegraphics[width=0.9\linewidth]{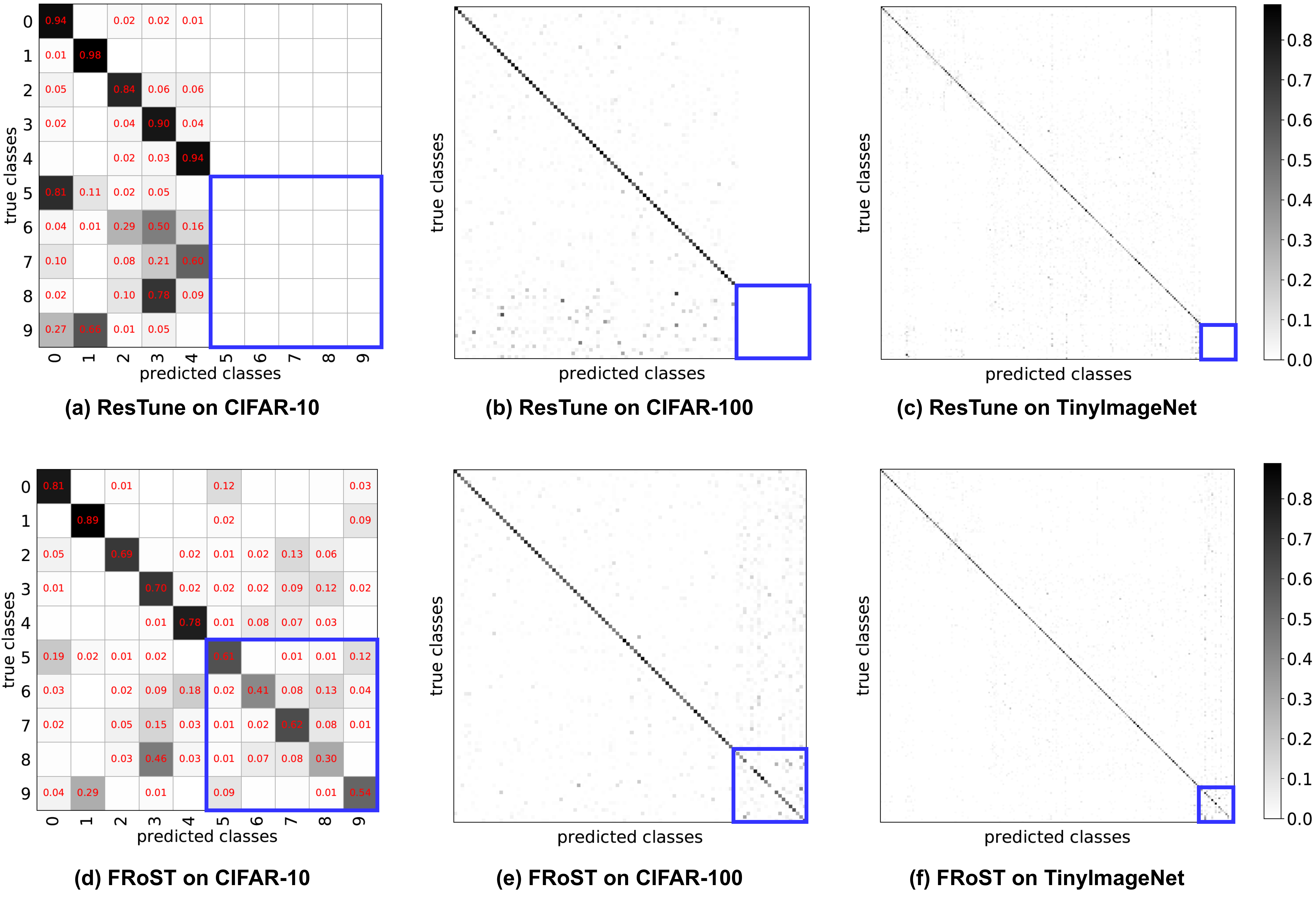}
    \caption{Comparison of the confusion matrices (CMs) for the \res (a)-(c) with our \ours (d)-(f) evaluated on all the data sets. The \bluebox\; denotes the part of the confusion matrix which corresponds to the new classes. While, the \res always predicts test samples from the new classes as belonging to the old classes, the CMs of \ours is mostly diagonal. This means that \ours can satisfactorily classify the new samples into the corresponding new classes. Digital zoom is recommended}
    \label{fig:cm-restuneVSfrost}
\end{figure}

In this section we compare the \res with our \ours under the \setting evaluation protocol. The comparison on all the data sets have been reported in the right halves of the Tab. \ref{tab:restune_comparison_cifar10}, \ref{tab:restune_comparison_cifar100}, and \ref{tab:restune_comparison_tiny}. Overall, under the \textbf{All} metric our proposed \ours consistently outperforms the \res. Given the \textbf{All} metric can be misleading, with the old classes dominating the performance over the new classes, we also compare the two methods using the \textbf{New} accuracy metric. As evident from the results of the \textbf{New} metric, we find that our joint classifier can satisfactorily classify the new classes in the task-agnostic evaluation setting, when compared to the \res. The \res results in a dismal performance of 0\%, meaning that \res is not suitable for discovering new classes in the task-agnostic setting. This can indeed be verified by visual inspection of the CMs reported in Fig. \ref{fig:cm-restuneVSfrost}. As can be observed, the CMs of \ours is more diagonal than that of \res, especially in the bottom part of CM which corresponds to the new classes. The well-behaved nature of \ours comes at the price of reduced \textbf{Old} accuracy. However, this is acceptable because the goal of \setting task is to simultaneously perform well in both old and new classes, unlike the highly skewed response in \res.

\subsection{Discussion}
\label{sec:app-discuss}

\begin{table}[!t]
\scriptsize	
\caption{The comparison of \ours with the state-of-the-art methods in the two-step \setting setting for the CIFAR-100 data set, where new classes arrive in two steps, instead of one. In the first step the metrics are: \textbf{New-1-J}: new classes performance with the joint-head (or concat-head); \textbf{New-1-N}: new classes performance from new-head. In the second step the metrics are: \textbf{New-1-J}: the performance of the previous 10 new classes; \textbf{New-2-J}: new classes performance}
\begin{center}
    \begin{tabular}{c|ccc|c|cccc|cc}
        \toprule
         \multirow{3}{*}{Methods}& \multicolumn{10}{c}{CIFAR-100} \\
         \cline{2-11}
         & \multicolumn{4}{c|}{First Step (80-10)} & \multicolumn{6}{c}{Second Step (80-10-10)} \\
         & \textbf{Old} & \textbf{New-1-J} & \textbf{All} & \textbf{New-1-N} & \textbf{Old} & \textbf{New-1-J} & \textbf{New-2-J} & \textbf{All} & \textbf{New-1-N} & \textbf{New-2-N}\\ 
         \midrule
         
         DTC\cite{Han2019LearningTD} & 61.0 & 0.0 &  54.2 & 51.5 & 50.0 & 0.0 & 0.0 & 40.0 & 42.6 & 58.9\\

         NCL\cite{zhong2021neighborhood} & \textbf{70.1} & 8.0 & \textbf{63.2} & 55.3 & \textbf{70.4} & 0.0 & 7.1 & \textbf{57.0} & 28.7 & 67.6 \\
         
         \hline
         ResTune\cite{liu2022residual} & 61.8 & 0.0 & 54.9 & \textbf{79.6} & 59.1 & 0.0 & 0.0 & 47.3 & 50.5 & 78.7 \\
         
         \textbf{FRoST} & 56.4 & \textbf{72.8} & 58.2 & 77.5 & 25.8 & \textbf{75.0} & \textbf{48.4} & 33.0 & \textbf{77.3} & \textbf{79.6} \\
         \bottomrule
    \end{tabular}
\end{center}
    \label{tab:multi-incd_cifar100}
\end{table}

We conjecture that the skewed predictions from the joint-head (or concat-head) of the \res is caused by the decoupled training of the separate classifier heads: old-head and new-head. Specifically, in \res, the old head is trained with the objective of LwF \cite{li2017learning}, whereas the new head is trained with a modified DTC \cite{Han2019LearningTD} objective. Due to lack of synergy between the two heads that receive gradients of different magnitudes, causes the predictions to be skewed (or biased) towards one of them. Contrarily, in our \ours the joint-head is always trained with cross-entropy (CE) loss. In more details, the CE loss is constructed from the feature-replay from the old class prototypes and the pseudo-labels from the novel-head corresponding to the new samples. Due to the usage of a homogeneous training objective for the joint classifier, the norm of the weights of the classifier is much balanced (see Fig. 4 in the main). Thus, our \ours can well predict both the old and the new classes, leading to a balanced performance and further justifying the validity of the proposed components. We believe that this insight is quite important and can be exploited in the future works.

\section{Additional Experiments for Two-Step \setting}
\label{sec:app-two-step}

In this section we present a detailed analysis of the two-step \setting setting, in addition to the experiments reported in the Sec. 4.3 of the main paper. We run a two-step \setting on CIFAR-100 (with 80 base classes) where in the first step we have 10 new classes. Subsequently, in the second \setting step we have another 10 new classes, resulting in a total of 90 old classes and 10 new classes. In the \textit{First Step} of Tab. \ref{tab:multi-incd_cifar100} we report the following metrics: {\em (i)} \textbf{Old} is the performance on the first 80 old classes; {\em (ii)} \textbf{New-1-J} is the performance on the 10 new classes seen during step 1; and {\em (iii)} \textbf{All} is the combined performance on all the classes seen until the end of the first step. All these metrics are computed with the joint-head for our \ours or using the concat-head for the baselines which do not support a joint-head. Note that we additionally report \textbf{New-1-N} that describes the performance of the new classes obtained using the new-head at the first step. Similarly, in the \textit{Second Step} when another 10 new classes are added we further report the \textbf{New-2-J} and \textbf{New-2-N} that deals with the performance of the newly added 10 classes from the joint-head and new-head, respectively. 

\begin{figure}[!ht]
    \centering
    \includegraphics[width=0.9\linewidth]{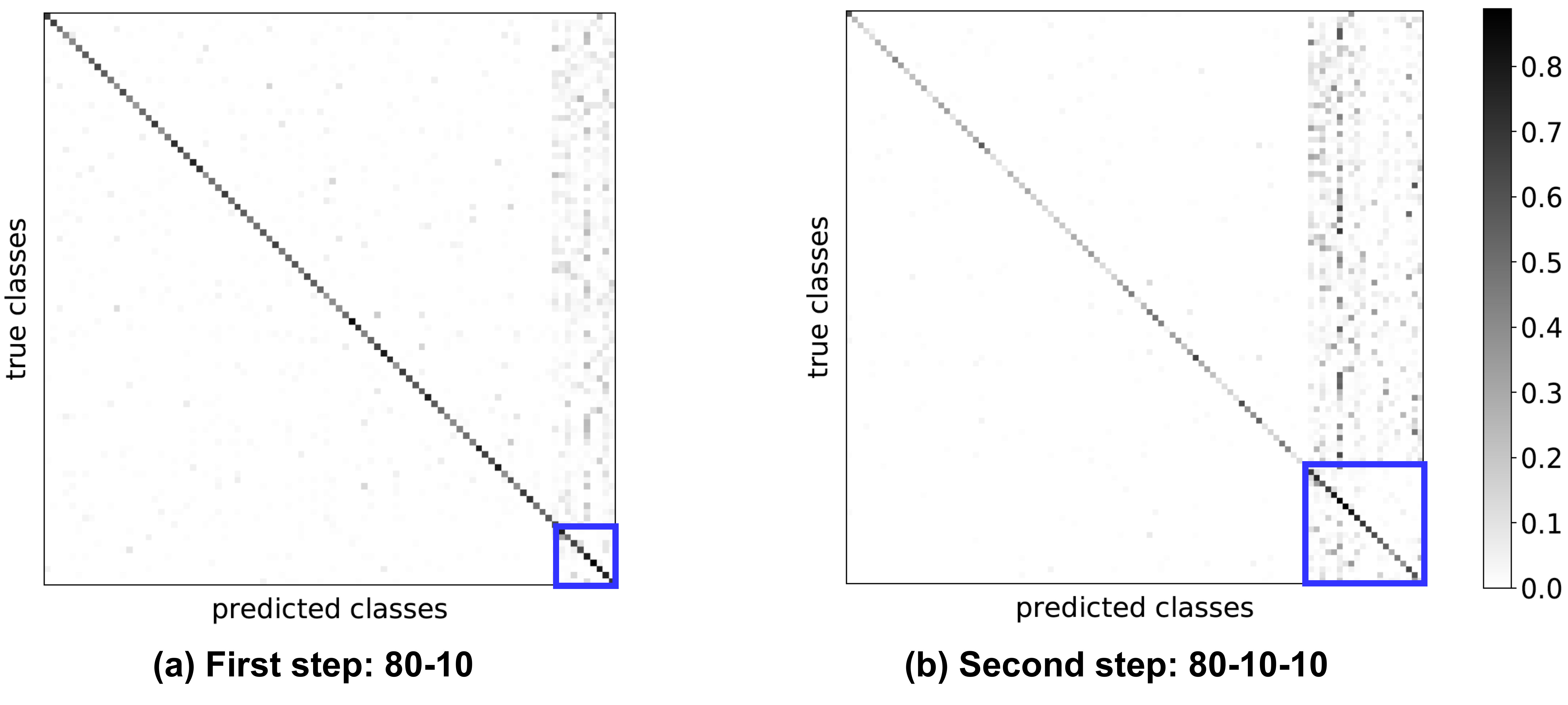}
    \caption{Confusion matrices (CMs) of the \ours in the sequential two-step \setting setting on CIFAR-100. (a) First step denotes the stage when we have 80 old classes and 10 new classes; and (b) Second step denotes the stage when we have an additional 10 new classes. The \bluebox\; denotes the part of the CMs which corresponds to the new classes seen after the supervised training on the old classes. The CMs are quasi-diagonal even after two steps and well-balanced over the old and the new classes}
    \label{fig:cm-two-step-frost}
\end{figure}

As can be seen from the Tab. \ref{tab:multi-incd_cifar100} our \ours achieves a well-balanced performance for both the old and the new classes in both the steps, in contrast to the \res, which fails to detect the new classes when evaluated with the concat-head. To prove that \res can discover the new classes when evaluated in a task-aware protocol, we report the performance of the new-head through the metrics \textbf{New-1-N} and \textbf{New-2-J}. Indeed, the task-aware new classes performance is at par with \ours in the first step, but experiences a drop in the second step. Thus \ours suffers from less forgetting as far as the first 10 new classes are concerned. We visually inspect the distribution in predictions through the CM of \ours in the Fig. \ref{fig:cm-two-step-frost} and observe that the CM is still quasi-diagonal, with some tendency to predict more the new classes in the second-step. Nevertheless, we improve over the baselines by a large margin when the overall performance is concerned.

\end{document}